\documentclass[
10pt, % Main document font size
a4paper, % Paper type, use 'letterpaper' for US Letter paper
oneside, % One page layout (no page indentation)
headinclude,footinclude, % Extra spacing for the header and footer
BCOR5mm, % Binding correction
]{scrartcl}

\usepackage[
nochapters, % Turn off chapters since this is an article        
beramono, % Use the Bera Mono font for monospaced text (\texttt)
eulermath,% Use the Euler font for mathematics
pdfspacing, % Makes use of pdftex’ letter spacing capabilities via the microtype package
dottedtoc % Dotted lines leading to the page numbers in the table of contents
]{classicthesis} % The layout is based on the Classic Thesis style

\usepackage{arsclassica} % Modifies the Classic Thesis package

\usepackage[T1]{fontenc} % Use 8-bit encoding that has 256 glyphs

\usepackage[utf8]{inputenc} % Required for including letters with accents

\usepackage{graphicx} % Required for including images
\graphicspath{{Figures/}} % Set the default folder for images

\usepackage{enumitem} % Required for manipulating the whitespace between and within lists

\usepackage{lipsum} % Used for inserting dummy 'Lorem ipsum' text into the template

\usepackage{subfig} % Required for creating figures with multiple parts (subfigures)

\usepackage{amsmath,amssymb,amsthm} % For including math equations, theorems, symbols, etc

\usepackage{varioref} % More descriptive referencing

\theoremstyle{definition} % Define theorem styles here based on the definition style (used for definitions and examples)

\theoremstyle{plain} % Define theorem styles here based on the plain style (used for theorems, lemmas, propositions)

\theoremstyle{remark} % Define theorem styles here based on the remark style (used for remarks and notes)

\hypersetup{
colorlinks=true, breaklinks=true, bookmarks=true,bookmarksnumbered,
urlcolor=webbrown, linkcolor=RoyalBlue, citecolor=webgreen, % Link colors
pdftitle={}, % PDF title
pdfauthor={\textcopyright}, % PDF Author
pdfsubject={}, % PDF Subject
pdfkeywords={}, % PDF Keywords
pdfcreator={pdfLaTeX}, % PDF Creator
pdfproducer={LaTeX with hyperref and ClassicThesis} % PDF producer
} % Include the structure.tex file which specified the document structure and layout

\usepackage{multirow}
\usepackage{array}
\newcolumntype{?}{!{\vrule width 2pt}}

\title{\normalfont\spacedallcaps{Convolutional Neural Networks for Automatic Detection of Artifacts from Independent Components Represented in Scalp Topographies of EEG Signals*}} % The article title

\author{\spacedlowsmallcaps{Giuseppe Placidi\textsuperscript{1}, Luigi Cinque\textsuperscript{2} \& Matteo Polsinelli \textsuperscript{1} }} % The article author(s) - author affiliations need to be specified in the AUTHOR AFFILIATIONS block

\date{} % An optional date to appear under the author(s)

\begin{document}

%----------------------------------------------------------------------------------------
%	HEADERS
%----------------------------------------------------------------------------------------

\renewcommand{\sectionmark}[1]{\markright{\spacedlowsmallcaps{#1}}} % The header for all pages (oneside) or for even pages (twoside)
\lehead{\mbox{\llap{\small\thepage\kern1em\color{halfgray} \vline}\color{halfgray}\hspace{0.5em}\rightmark\hfil}} % The header style

\pagestyle{scrheadings} % Enable the headers specified in this block

%----------------------------------------------------------------------------------------
%	TABLE OF CONTENTS & LISTS OF FIGURES AND TABLES
%----------------------------------------------------------------------------------------

\maketitle % Print the title/author/date block

\setcounter{tocdepth}{2} % Set the depth of the table of contents to show sections and subsections only

%\tableofcontents % Print the table of contents

%\listoffigures % Print the list of figures

%\listoftables % Print the list of tables

%----------------------------------------------------------------------------------------
%	ABSTRACT
%----------------------------------------------------------------------------------------

\section*{Abstract} % This section will not appear in the table of contents due to the star (\section*)
Electroencephalography (EEG) measures the electrical brain activity in 
real-time by using sensors placed on the scalp. Artifacts, due to eye movements and 
blink, muscular/cardiac activity and generic electrical 
disturbances, have to be recognized and eliminated to allow a correct interpretation of the useful brain signals (UBS) of EEG. Independent 
Component Analysis (ICA) is effective to split the signal into independent components (ICs)
whose re-projections on 2D scalp topographies (images), also called topoplots, allow to 
recognize/separate artifacts and by
UBS. Until now, IC topoplot analysis, a gold standard in EEG, has been carried 
on visually by human experts and, hence, not usable in automatic, fast-response EEG. We present 
a completely automatic and effective framework for EEG artifact 
recognition by IC topoplots, based on 2D Convolutional Neural
Networks (CNNs), capable to divide topoplots in 4 classes: 3 types of 
artifacts and UBS.
The framework setup is described and results are presented, 
discussed and compared with those obtained by other competitive strategies. Experiments, carried on public EEG datasets, have shown an overall accuracy of above 98\%, employing 1.4 sec on a standard PC to classify 32 
topoplots, that is to drive an EEG system of 32 sensors. Though not real-time, the proposed framework is 
efficient enough to be used in fast-response EEG-based Brain-Computer Interfaces 
(BCI) and faster than other automatic methods based on ICs.

%----------------------------------------------------------------------------------------
%	AUTHOR AFFILIATIONS
%----------------------------------------------------------------------------------------

\let\thefootnote\relax\footnotetext{\textsuperscript{1} \textit{A2VI Lab, Dept. of Life, Health
and Environmental Sciences, University of L’Aquila, Via Vetoio, L’Aquila 67100, Italy}}

\let\thefootnote\relax\footnotetext{\textsuperscript{2} \textit{Dept. Computer Science, Via Salaria, Sapienza University, Rome, Italy}}

\let\thefootnote\relax\footnotetext{* \textit{Manuscript submitted to Computer Methods and Programs in Biomedicine}}

%----------------------------------------------------------------------------------------

\newpage % Start the article content on the second page, remove this if you have a longer abstract that goes onto the second page

%----------------------------------------------------------------------------------------
%	INTRODUCTION
%----------------------------------------------------------------------------------------

\section{Introduction}

EEG measures, through electrodes placed on the scalp, the neural activity with an excellent temporal resolution. Optimal temporal resolution and low invasiveness make EEG particularly suitable for fast response Brain Computer Interfaces (BCI)~\cite{suk2012novel, cecotti2010convolutional, Placidi2015}. Extraneous signals produced by eye movements and blink, muscular spasms, cardiac activity and generic inteferences~\cite{Uriguen2015, noureddin2011online} can obscure UBS, since skull and scalp (including muscles) are in between brain and sensors. 

Blink and eye movements produce electrooculography (EOG) artifacts which are mainly recorded by frontal sensors and propagate across the scalp~\cite{Joyce2004}. Three categories of EOG exist: eye blink (BEOG), vertical (VEOG) and horizontal(HEOG) EOG. EOG have often much higher amplitude than UBS and frequencies in the range 10-40 Hz where also UBS are present.

Cardiac activity produces electrocardiography (ECG) artifacts~\cite{Lin2014}. Most of ECG effects can be reduced by subtracting the signal of a peripheral sensor (usually placed on one of the ears) from those of sensors on the scalp~\cite{noureddin2011online}. Residual ECG artifacts, when present, are lower than brain signals and often tolerated. 

Cranial muscles produce electromyogram (EMG) artifacts. The main feature of EMG is the wide spectral distribution with maximum power in the range of 15-30 Hz where also UBS insist~\cite{Shibas1999}.

Finally, generic discontinuities are generated by impedance fluctuations or electric/electronic interferences (IF)~\cite{Mognon2011}. They affect single sensors with large fluctuations in amplitude. 

Artifacts power could be much higher than that of UBS and propagate to large regions~\cite{Joyce2004, Mognon2011,Delorme2007}: UBS could be completely obscured, resulting in a wrong interpretation of EEG, if an effective pre–processing strategy is not employed. As discussed above, artifacts could be confused with UBS in frequencies and alternatives to time-frequency analysis have to be used. Besides coding, also a good spatial representation of the sources generating signals is important because information such as shape and localization are fundamental to discover the signal nature~\cite{zhang2019making}. For example, the reprojection of a signal source on the scalp allows understand whether it is an EOG artifact or an UBS. Fortunately, EEG signals can be viewed as mixtures of linear independent source components, some mainly due to artifacts and others to UBS~\cite{Jung2000163}. 
%This means that it is possible, for a human expert, to detect artifact components if a visual representation can be defined for them. 

\begin{figure*}[!t]
	\centering
	\includegraphics[width=0.35\linewidth]{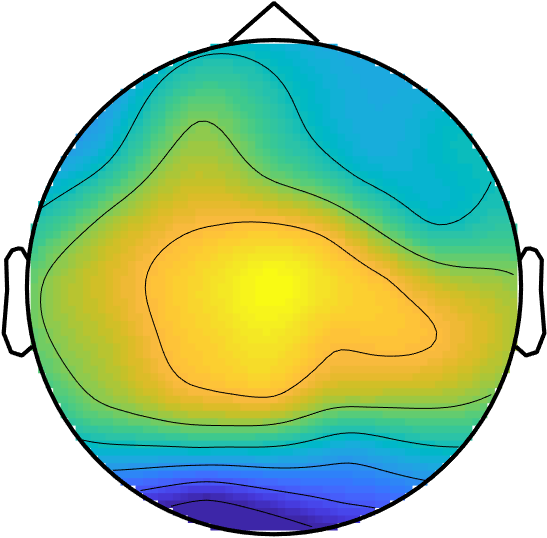}
	% where an .eps filename suffix will be assumed under latex, 
	% and a .pdf suffix will be assumed for pdflatex; or what has been declared
	% via \DeclareGraphicsExtensions.
	\caption{An example of topoplot representation of an Independent Component. Colours are indicative of the underlying activation of different brain regions: blue represents low activation, yellow correspond to high activation.}
	\label{fig:topoplot}
\end{figure*}

An effective method to retrieve EEG source components is Independent Component Analysis (ICA)~\cite{Delorme2007}. Since EEG is usually measured on $n$ channels, for each temporal slot a maximum set of $n$ components could be calculated by ICA. Each component is defined by an array of weights representing the magnitude of each sensor related to that component. Weights can be interpolated in 3D (by using the spatial localization of the sensors on the scalp) and projected on a 2D scalp map called topoplot~\cite{Raduntz2015} to allow spatial localization of the component (Fig~\ref{fig:topoplot}). 

Visual inspection of the represented scalp topography allows a human expert, on the basis of the component localization, to understand its nature  and to produce a classification~\cite{Raduntz2015}. In fact though the consultation of other information, such as power spectrum density, could help to avoid that some UBS is confused with artifacts, they are seldom used due to the huge increment of the time needed for visual classification (to a trained expert, visual inspection of one topoplot requires about 5-7 secs: the addition of power spectrum density analysis could double this time).
Human visual inspection is effective and widely used for effective, though off-line, artifacts removing: in fact, it is supported in most of the software for EEG signals processing such as EEGLAB~\cite{Delorme2004} and FieldTrip~\cite{Oostenveld2011}. 

In the present paper, we propose a completely automatic and fast method to perform pre-processing of EEG signals by recognizing artifacts from topoplots through a CNN-based approach emulating the human visual interpretation process. 

Main contributions of the manuscript are: to define, train, validate and test a CNN-based framework for effective EEG artifacts recognition by topoplot representation of Independent Components;  to allow artifact recognition in an efficient way to be used for fast-response EEG; to obtain a general purposes and scalable framework, independent of the number and disposition of sensors, that is capable to deal with any kind of known artifacts andthat could be easily adapted to future artifact patterns by adding new elements without retrain the whole system; to define a public dataset of labelled topoplots, for testing the algorithms for artifacts recognition, based on data collected by DEAP~\cite{Koelstra2011deap}, a web-based multi-centres dataset of EEG signals. 

The manuscript is organized as follows: Section 2 reviews related works on EEG artifacts reduction; Section 3 describes the proposed  framework; Section 4 reports the experimental dataset, the training procedure, the experimental results, their discussion and comparison with other competitive methods; Section 5 concludes the work and discusses future improvements and applications. The trained proposed framework and the labelled dataset of topoplots, with related documentation, can be found at \url{https://drive.google.com/open?id=1NRibJzEYInZm28sn4Esx7bDc2isNz3lM}. 

\section{Related Works}

Several techniques have been designed and used for EEG artifacts detection~\cite{Islam2016, Mannan2018, Uriguen2015}, roughly grouped in the following categories: Regression Methods; Filtering Algorithms;
Wavelet Transform; Empirical mode decomposition; Blind Source Separation.

Regression Methods assume that artifacts are measured through dedicated channels~\cite{VandenBerg-Lenssen1994, Mannan2018}. Measurements are necessary to estimate the propagation coefficients that can be subtracted to brain signals. Though these methods have good computational performance, they have two major drawbacks: one or more reference channels are needed, a disadvantage for fast-response EEG~\cite{Minguillon2017}; regressing out artifacts involves subtracting also relevant UBS~\cite{Mognon2011}. 

Filtering include several approaches: between them, adaptive filtering is the most used~\cite{Romero2008}. This method assumes that UBS are uncorrelated from artifacts and requires that a dedicated channel is used to measure artifacts which are subtracted to the signal. Filtering has the same limitations of regression methods~\cite{Sweeney2012}. 

Unlike Regression Methods and Filtering, Wavelet Transform (WT) does not require reference signals. WT transforms the EEG signal from time-domain to time-frequency domain. Once the signals are decomposed, the calculated coefficients are thresholded prior being inverse transformed. The main issue is that WT is unable to completely remove artifacts that spectrally overlap with UBS~\cite{Mannan2018, Uriguen2015} or that are too specific ~\cite{acharyya2018low}.

Empirical mode decomposition (EMD) performs operations that decompose the signal into amplitude and frequency modulated basis functions, in the time domain (Intrinsic Mode Functions, IMFs). This method is useful for analyzing non-linear and non-stationary natural signals, has good performance for artifact removal but is seldom used due to its computational complexity~\cite{Uriguen2015} that makes it very slow and, hence, not adapted for fast response EEG.

\begin{figure*}[!t]
	\centering
	\includegraphics[width=\linewidth]{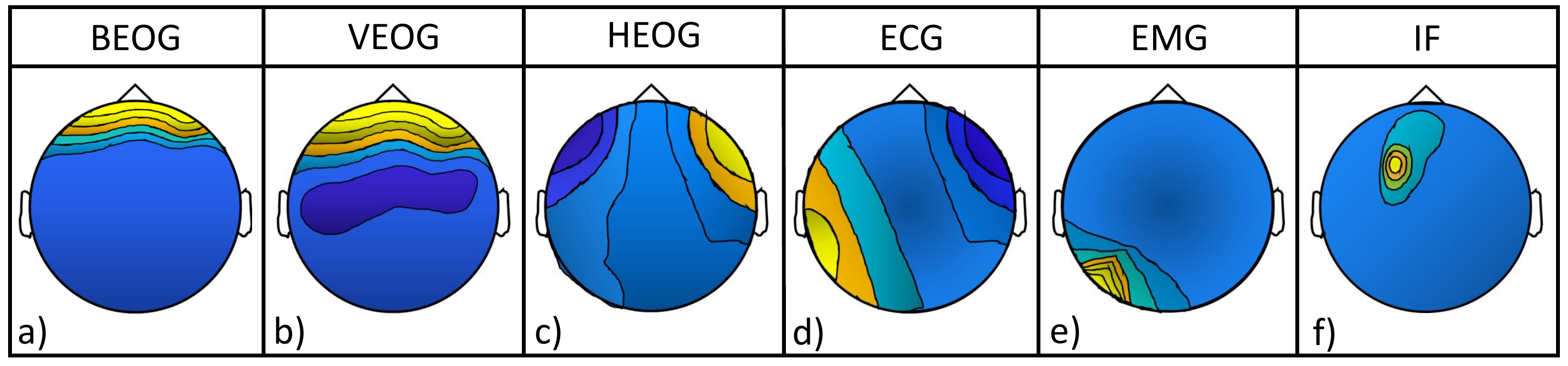}
	\caption{Topoplot representation of ICA components related to artifacts. BOEG (a) and VEOG (b) have similar shapes. The same occurs between HEOG (c) and ECG (d) and between EMG (e) and IF (f). EOG and ECG have well-defined scalp locations; EMG and IF are isolated peaks that can occur in different scalp positions.}
	\label{fig:artefatti_topoplots}
\end{figure*}

Blind Source Separation does not require prior information and/or reference signals. Two algorithms have been used over time: Principal Component Analysis (PCA) and ICA. One of the first attempt to use PCA was in 1991~\cite{Berg1991} but in 1997 it was demonstrated that PCA can not completely separate artifacts by UBS~\cite{Lagerlund1997}. Indeed PCA transforms time-domain observations of correlated variables into a set of linearly uncorrelated variables using an orthogonal transformation: when UBS and artifacts are not orthogonal and their amplitude are comparable, PCA fails to separate the corresponding components~\cite{Mannan2018}. ICA overcomes these limitations, even if UBS and artifacts have comparable amplitudes~\cite{Delorme2007,Zhao2014109} and, for this reason, it is considered the most effective approach to analyze EEG signals~\cite{Urrestarazu2004, Vigario2000, Uriguen2015}. However, it is necessary to represent Independent Components (IC) in 2D scalp topographies and to consider their resulting shape, in order to establish whether they allow to artifacts or to UBS. Fig~\ref{fig:artefatti_topoplots} shows characteristic topoplot shapes of the artifacts presented above~\cite{Mognon2011, Raduntz2015, Raduntz2017}.

As can be seen, BEOG concentrates its power on the frontal region of the scalp (Fig~\ref{fig:artefatti_topoplots}.a) as VEOG (Fig~\ref{fig:artefatti_topoplots}.b), though VEOG  spreads across the scalp more than BEOG. For their similarity both in pattern and in significance they can be grouped into a single class. 

In the case of HEOG, two peaks of opposite sign are positioned around the nose (Fig~\ref{fig:artefatti_topoplots}.c). Similar to HEOG, ECG (Fig~\ref{fig:artefatti_topoplots}.d) is composed by two peaks of opposite sign localized on the scalp borders, around the ears (ECG differs from HEOG just for a different orientation). The similarity between HEOG and ECG suggests their inclusion into the same class (the recognition between them, out the scope of the present manuscript, could be based on the orientation of the peaks).

EMG and IF are isolated peaks, the first usually located on the border of the scalp close to neck and face where muscular activity is pronounced (Fig~\ref{fig:artefatti_topoplots}.e), while the second frequently located in the middle of the head (Fig~\ref{fig:artefatti_topoplots}.f). Due to their similar shape, EMG and IF are included into the same class, though their nature is very different (EMG are muscular spasms and IF are electrical disturbances). The distinction between a channel failure and a muscular spasm is difficult also for a human expert: position and frequency of occurrence increase the probability of one with respect to the other but, for our purposes, both need to be discarded. 

As specified above, the role of IC scalp topography is fundamental for a human expert to identify artifacts. However, EEG measurements could consist of several channels, several temporal windows (trials) also divided in sub-windows (sub-trials) and the amount of ICs, that is of topoplots, could easily reach hundreds of thousands. These numbers make visual inspection and recognition impossible, especially when fast response is required. 

In~\cite{Islam2016}, of 46 works reviewed, just 3 of them use ICs~\cite{James2003, Winkler2014, Winkler2014a}, though their common issue is that they were all designed just to deal with specific artifacts and not capable to remove all types of artifacts.
A recently proposed, ICs-based, automatic method ~\cite{Pion2019} is a classifier that uses power spectrum density (PSD) and autocorrelation to support and/or surrogate the loss of information occurring from the usage of low resolution (32x32) IC topographies. Low resolution in IC topoplots was necessary to gain efficiency (the generation of topoplots is computationally expensive). Though effective, the resulting classification strategy is particularly difficult to train due to the difficulty to obtain labelled data from human experts. In fact, long time is required for manual classification and for the definition of threshold parameters for PSD and autocorrelation. Furthermore, the strategy is not very adaptable and generalizable to different scenarios (i.e. when data with different signal-to-noise ratio are used, the system has to be re-trained).         

To the best of our knowledge, none of the previous strategies satisfies the following conditions at the same time: to recognize all types of artifacts actually known with good accuracy just from IC topoplots; to be easy and easily trained; to be independent of the signal quality; to be independent of the number of EEG sensors; to be scalable with respect to newly discovered sources of artifacts; to be efficient enough for fast-response EEG applications, such as BCIs.

\section{The Overall Framework}
\begin{figure*}
	\begin{center}
		\includegraphics[width=\linewidth]{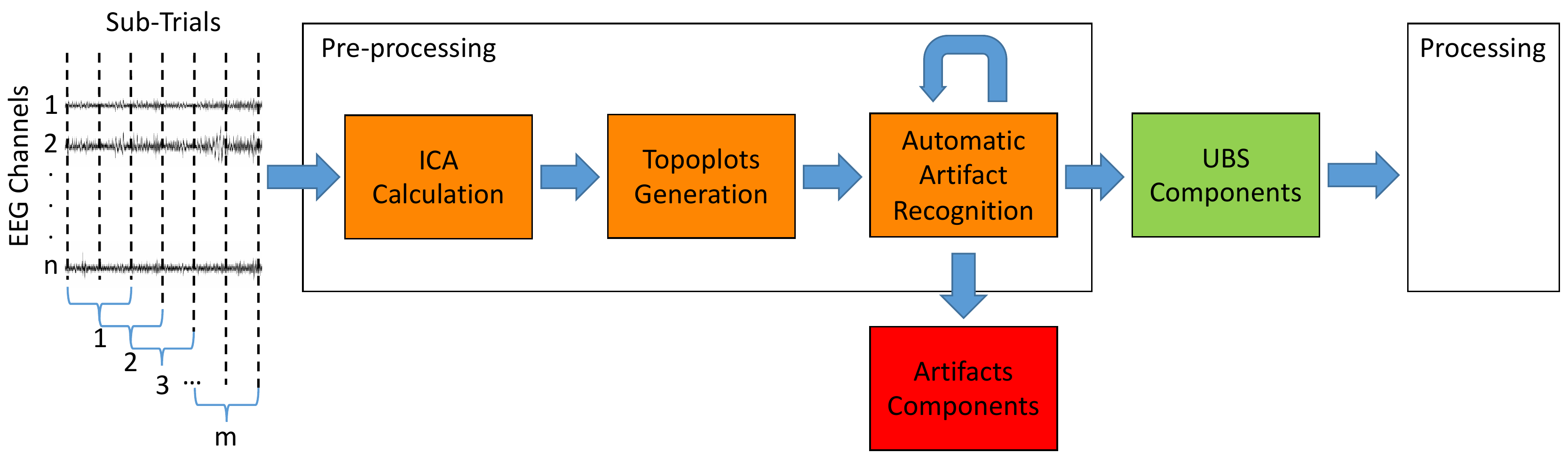}
	\end{center}
	\caption{
		Pre-processing pipeline. It includes: ICA calculation of EEG partially overlapped temporal sub-trials (minimal duration, defined by curly brackets on the left, to allow ICA convergence); generation of the resulting topoplots; automatic artifact recognition by topoplot images. The resulting UBS are passed to the processing step. 
	}
	\label{fig:pipeline}
\end{figure*}

In what follows we aim at replacing the role of the human expert and his visual inspection of IC topoplots with a completely automatic strategy, based on the analysis of the EEG signal separated in sub-trials, that is to on partially overlapped pieces of EEG signals along time (to take into account for the transient nature of the artifacts), as reported in Fig~\ref{fig:pipeline}.

Recently, Convolutional Neural Networks have revolutionated computer vision, in particular regarding recognition of objects from images~\cite{al2017review,song2018deep,zhang2019making}.
The framework we propose is based on CNNs, since we aim at detecting artifacts from EEG ICs topoplot images and that CNNs have been successfully used in several EEG classification studies ~\cite{cecotti2010convolutional,askari2018modeling,tang2019motor,Pion2019,gao2020automatic}.
Our objective is to allow the separation of artifacts by UBS and to classify artifacts into the three classes defined above. This last choice would not be justified by a simple recognition/elimination process: our aim is to separate artifacts in classes to allow the assumption of future decision on them, in real-time. For example, if an artifact of the class EMG/IF occurs frequently, it can be argued that it is due to a sensor failure (IF) more than to a muscular spasm (EMG): in that case, the definitive elimination of data from the sensor could be more convenient and efficient than continuously discovering/eliminating the artifacts it generates. This type of decision could be of paramount importance in BCIs.
Other objectives are: to stress on generality and scalability by foreseeing the treatment of topoplot patterns regarding future artifacts without re-defining the overall structure and, most important, without re-train the whole system; to reduce the training dataset thus reducing the overload for human experts in recognizing and labelling data for training (a very long and boring task).

\begin{figure*}
	\begin{center}
		\includegraphics[width=\linewidth]{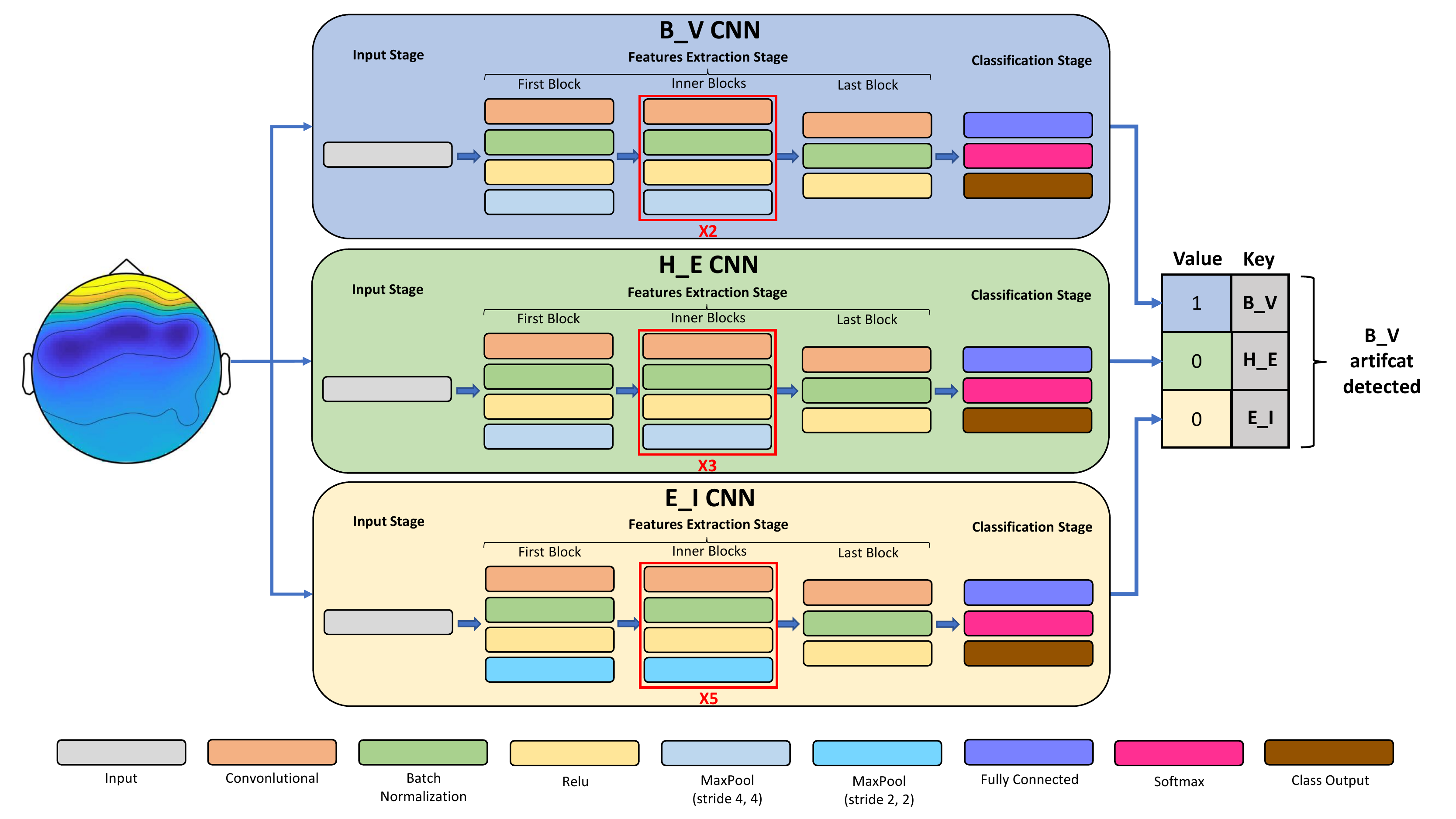}
	\end{center}
	\caption{The proposed framework architecture. The disposition of the 3 CNNs reflects the classes of artifacts (one for each class). The current topoplot is passed to the three CNNs separately. In the proposed case, an example, a classical BEOG artifact is the input of the framework: the resulting output should indicate its recognition by the first CNN (1) and a refutation by the other two (0). The Input and the Classification Stages have the same design in all the CNNs. Regarding the Features Extraction Stage, B\_V CNN contains 2 Inner Blocks, H\_E CNN 3 Inner Blocks and E\_I 5 Inner Blocks.}
	\label{fig:CNN_completeArchitecture}
\end{figure*}

To pursue the previous objectives, we propose the architecture in Fig~\ref{fig:CNN_completeArchitecture}, composed by 3 parallel CNNs of the type described in~\cite{Lecun1989, lecun1998gradient}. In fact, because of their common patterns, BEOG and VEOG are in a single CNN (B$\_$V CNN), as HEOG and ECG (H$\_$E CNN) and EMG and IF (E$\_$I CNN). The fact that we have grouped different artifacts into a single CNN is threefold: 1) artifacts allowing to the grouped sub-classes are often difficult to be recognized each-other also by a human expert; 2) grouped artifacts share the same treatment 3) computational efficiency is improved. 
Besides the advantages of grouping, the proposed framework has some advantages with respect to extreme grouping, i.e. the usage of a single CNN: speed-up training, lowering datasets for training, separating treatment strategies, speed-up convergence and increased generality. The previous advantages can be mostly explained by the fact that each CNN of the framework can be trained separately.

\subsection{Framework Design}

CNNs are based on feed-forward artificial neural networks (ANNs). A CNN consists of input and output layers and multiple hidden layers for feature extraction, including convolutional layers.  The main advantages of CNNs with respect to classical ANNs is that neurons in one layer do not connect to all the neurons in the next layer but only to a small subset. The three CNNs used therein are structured as shown in Fig \ref{fig:CNN_completeArchitecture}.

Each CNN is organized in 3 stages: an Input Stage, a Classification Stage, interleaved by a Features Extraction stage. The Input Stage is composed only by an Input Layer. The Classification Stage is composed by a Fully Connected layer (of dimension 2), a Softmax layer and a Classification layer. Input and Classification Stages are the same for all the CNNs. The Features Extraction Stage extracts different features for each class of artifacts (geometrical properties, position and orientation inside the topoplot, intensity, etc) and, for this reason, is specific for each CNN and organized into "Blocks". Each Block contains a Convolutional layer, a BatchNorm layer, a Relu layer and a MaxPool layer, except for the last Block where the MaxPool layer is absent. Since the First Block is executed separately on each of the three channels of the RGB topoplot, the three output are summed before being passed to the next Blocks. The number of Inner Blocks, of filters of the Convolutional Layers and the stride of the MaxPool Layers are adjusted for each CNN during multiple training phases until each CNN reaches its proper best accuracy.

In the B\_V CNN, the Features Extraction Stage is composed by 2 Inner Blocks, the Convolutional layers use 8, 16, 32 and 64 filters and the MaxPool layers have dimension 2x2, stride [4,4] and padding 0.

H\_E CNN contains 3 Inner Blocks and the Convolutional layers use 8, 16, 32, 64 and 128 filters (the MaxPool layers have the same dimension of B\_V CNN).

Finally, in I\_F CNN, the Features Extraction Stage is composed by 5 Inner Blocks and 8, 16, 32, 64, 128, 256, 256 filters for the Convolution layers. In this case, the MaxPool layers have still dimension 2x2 but stride is [2, 2] (this resembles the fact that Impedence and EMG artifacts are composed by a few neighboring pixels inside the topoplot). Feature Extraction Stage is specific for each class because:

\begin{itemize}
	\item patterns of B\_V  are well defined and localized (low complexity);
	\item patterns of H\_E  are less defined than in B\_V (medium complexity);
	\item patterns of E\_I are neither well localized nor well defined (high complexity). 
\end{itemize}

\section{Experimental evaluation}

In what follows we present the experimental dataset collection, the procedure used to train, validate and test the system and the experimental results. Besides efficacy, also efficiency is considered.

The framework was implemented in Matlab (The MathWorks Inc., https://mathworks.com/) on a PC with Intel Core I7-6700, 32GB of RAM and Nvidia GeForce GTX 1080. 

\subsection{Experimental Dataset}

The experimental dataset consisted of EEG  signals collected by the DEAP  dataset~\cite{Koelstra2011deap},  a  public multi-center database containing a collection of EEG signals of negative and  positive emotional states by 32 participants (16 men and 16  women,  aged  between  19  and  37,  average:  26.9)  recorded  while  watching  40,  one-minute  long,  music  videos  on  different  arguments.  In  particular,  participants  watched music videos and rated each of them in terms of   arousal,   valence,   like/dislike,   dominance,   and   familiarity. EEG   signals,   sampled  at  512  Hz,  were  recorded  on  the  following 32 positions (according to the international 10-20 positioning system~\cite{nuwer1998ifcn}): Fp1, AF3, F3, F7, FC5, FC1, C3, T7, CP5, CP1, P3, P7, PO3, O1, Oz, Pz, Fp2, AF4, Fz, F4, F8, FC6, FC2, Cz, C4, T8, CP6,  CP2,  P4,  P8,  PO4,  and  O2. For more information on DEAP please refer to~\cite{Koelstra2011deap}. For our porposes, DEAP raw data, after filtering with a notch filter~\cite{Leske2019} at 50Hz and 60Hz to suppress power-line interferences, where used. 
Data were divided in sub-trials, 8 seconds long overlapped by 4 seconds (4 seconds of new signal joint to 4 seconds of "past" signal) and used to generate ICA components and the corresponding topoplots. The usage of "past" signal to support the current signal was made to fit both the following apparently contrasting requirements:
\begin{enumerate}
	\item to take a sufficiently long signal to ensure ICA convergence; 
	\item to take a sufficiently short new signal window for fast response EEG applications.
\end{enumerate} 
This choice also represents a good compromise to cancel out transient artifacts without compromising UBS components (the analysis of longer time sequences would average the artifacts contribution with UBS).

\begin{table}[t]
	\begin{center}
		\resizebox{\textwidth}{!}{%
			\begin{tabular}{|c|c|c|}
				\hline
				\textbf{B\_V CNN}         & \textbf{H\_E CNN}          & \textbf{E\_I CNN}            \\ \hline
				1341 (B\_V)           & 398 (H\_V)             & 1592 (E\_I)                 \\ \hline
				5020 (H\_E + E\_I + UBS) & 4823 (B\_V + E\_I +UBS) & 6044 (B\_V + H\_E +UBS) \\ \hline
		\end{tabular}}
	\end{center}
	\caption{Composition of the datasets used to train each CNN.}
	\label{tab:table1}
\end{table}

Regarding the pre-processing phase (the only we were interested), ICA was performed on each sub-trial and a maximum of 32 components, at most one for each channel, was calculated. For each IC, a topoplot was generated through FieldTrip and managed as a 134x136 RGB image of fixed position and orientation. 
This resolution represented a good trade-off between high spatial accuracy and execution time reduction. Another constraint was that topoplots were represented in 64 colors Parula color palette. This palette is commonly used in problems solved by CNNs (\cite{liu2018method},\cite{ito2018application},\cite{siddharth2019utilizing}). Since our framework was trained on Parula topoplots, it is capable to deal just images in Parula: transformations are necessary, before the application of the method, if other palettes are used.
The number of obtained topoplots was 992800. From this dataset, the images to be used for CNNs training/validation/test were extracted as described below.

We could not use data augmentation strategies (changes of orientation, scaling, translation and brightness) because:

\begin{enumerate}
	\item the orientation of each topoplot is fixed and rotations would change its significance and, hence, would be unjustified;
	
	\item scaling and translation would create redundancy because the interpretation of a topoplot is always referred to the external silhouette of the head;
	
	\item brightness augmentation would be wrong because IC topoplots are generated with fixed colormaps of fixed brightness scale, as discussed above. 
	
\end{enumerate}

However, due to the availability of a huge dataset, the absence of data augmentation will not be a problem for the learning process, as will be clarified in Result and Discussion section.

\subsection{Training}

DEAP data of subjects 1-8 were manually classified and labelled independently by 5 human experts into 4 different categories: BEOG $\cup$ VEOG (B\_V), HEOG $\cup$ ECG (H\_E), EMG $\cup$ IF (E\_I) and UBS. A consensus dataset was obtained by considering each topoplot allowing to the most voted class. In case of equity (the case in which the most voted classes were 2 with 2 votes each), the resulting topoplot was discarded. The consensus dataset agreed at 95.7\% with all the human experts (agreement between consensus and each of the experts was between 97.2\% and 99.1\%).

From the labelled consensus, all the artifacts where extracted and just a subset of UBS was randomly selected. The composition of the training sets is illustrated in Table \ref{tab:table1}.

\begin{figure}
	\begin{center}
		\includegraphics[width=0.65\linewidth]{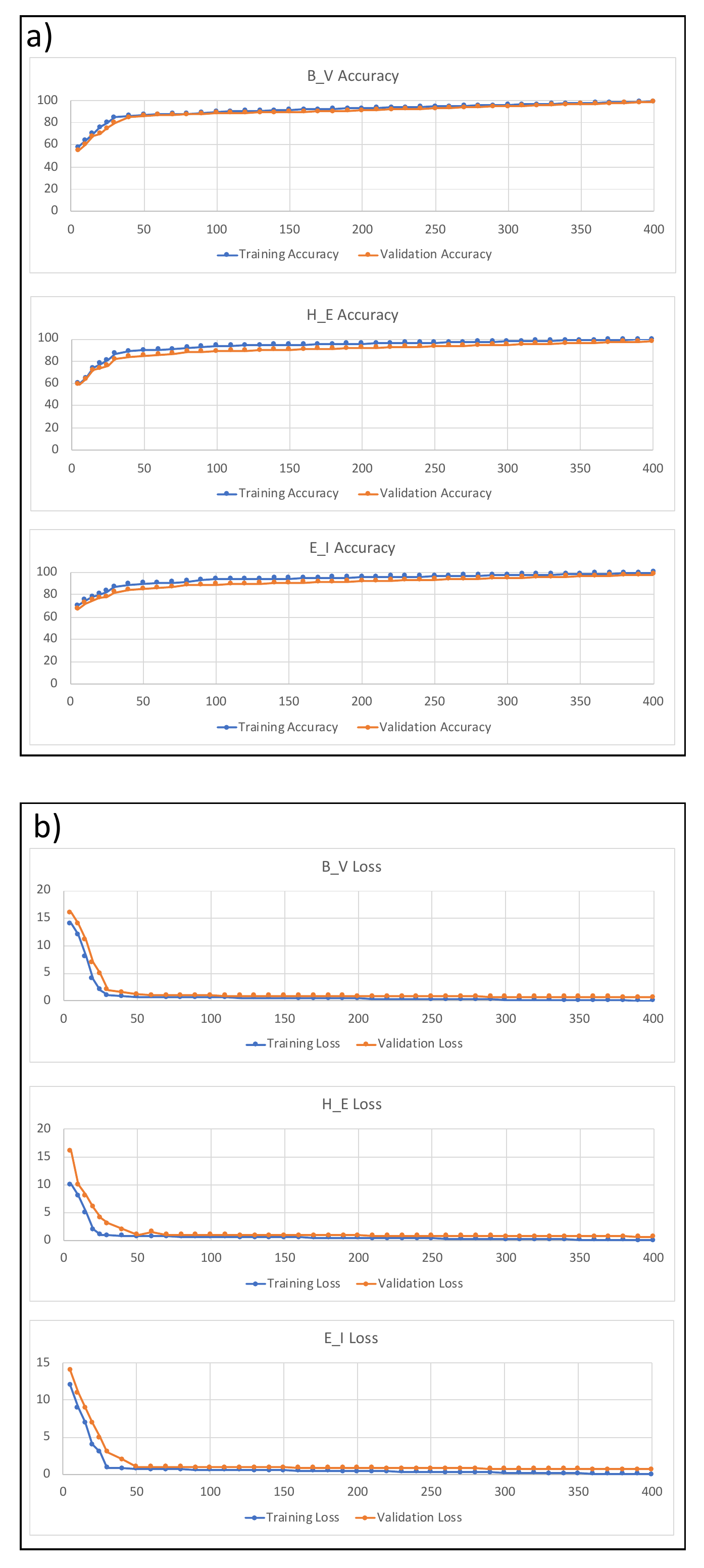}
	\end{center}
	\caption{Accuracy \% (a) and Loss values (b), with respect to epochs, for the three CNNs: B\_V, H\_E and E\_I, respectively. Blue is used for training and red is used for validation.}
	\label{fig:training_graphics_cnn}
\end{figure}

\begin{figure}[t]
	\begin{center}
		\includegraphics[width=0.4\textwidth]{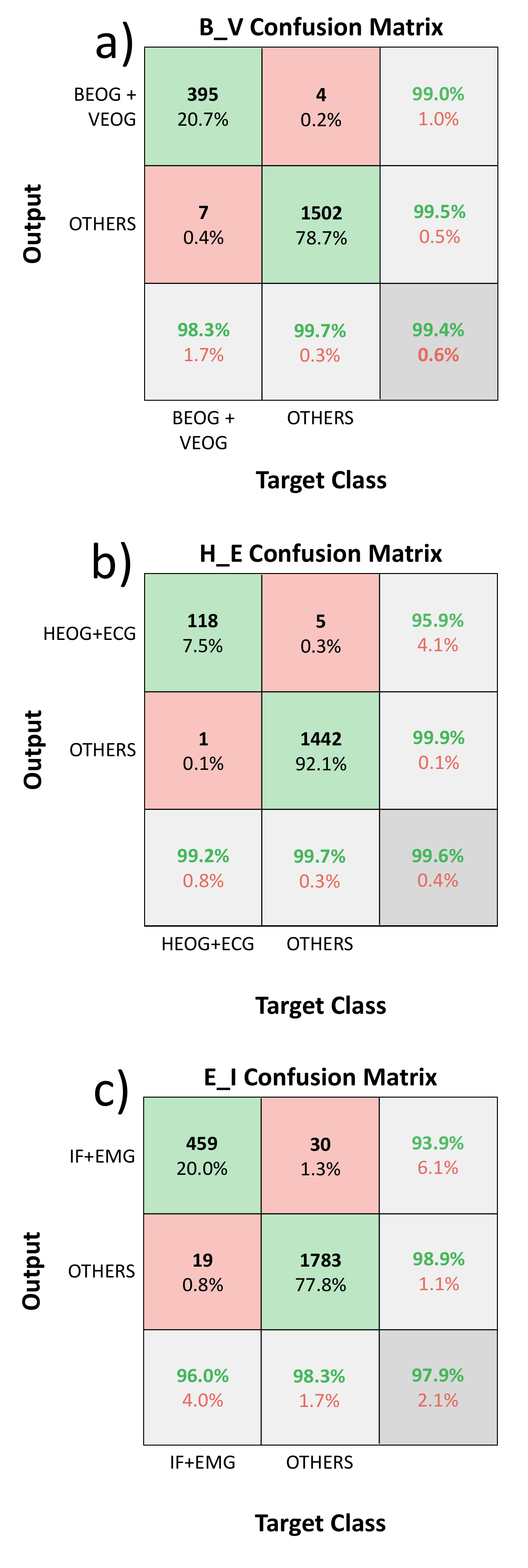}
	\end{center}
	\caption{Confusion matrices for the fifth repetition of the validation process of: B\_V (a); H\_E (b); E\_I (c). Each cell contains the absolute value and the corresponding \% with respect to the number of elements.}
	\label{fig:matrix}
\end{figure}

In particular, for each CNN, the training set was organized by separating the resulting topoplots in two classes: the first class containing the artifacts to be recognized by the CNN and the second containing all the others (other artifacts + UBS), as reported in Table \ref{tab:table1}. In this way, each CNN was trained to recognize its proper artifacts and to separate them from the other signals (the number of artifacts into each of the three classes approximately reflects the frequency of their occurrence in real scenarios). 
%\begin{table*}
%\begin{center}
%\begin{tabular}{ccccl}
%\cline{1-4}
%\multicolumn{1}{|c|}{\textbf{Class}} 	& \multicolumn{1}{c|}{\textbf{B\_V CNN}}    		& \multicolumn{1}{c|}{\textbf{H\_E CNN}}          		& \multicolumn{1}{c|}{\textbf{E\_I CNN}}           				&  \\ \cline{1-4}
%\multicolumn{1}{|c|}{Class 1}       		& \multicolumn{1}{c|}{1341 (BEOG/VEOG)}			& \multicolumn{1}{c|}{398 (HEOG/ECG)}             			& \multicolumn{1}{c|}{1592 (EMG/IF)}                 					&  \\ \cline{1-4}
%\multicolumn{1}{|c|}{Class 2}        	& \multicolumn{1}{c|}{5020 (HEOG/ECG+EMG/IF+UBS)}		& \multicolumn{1}{c|}{4823 (BEOG/VEOG+EMG/IF+UBS)} 	& \multicolumn{1}{c|}{6044 (BEOG/VEOG+HEOG/ECG+UBS)} 			&  \\ \cline{1-4}
%\multicolumn{1}{l}{}                		& \multicolumn{1}{l}{}                   				& \multicolumn{1}{l}{}                        				& \multicolumn{1}{l}{}                         						& 
%\end{tabular}
%\end{center}
%\caption{Composition of the datasets used for training of each CNN.}
%\label{tab:table1}
%\end{table*}

The training dataset of each CNN was divided 70\% for training and 30\% for validation. A Stocastic Gradient Descendend with momentum (momentum 0.9) was the optimization algorithm. The gradient threshold method was the L$_2$-norm and the max number of epochs was fixed to 400, though  all the CNNs fully converged well below 100 epochs (Fig~\ref{fig:training_graphics_cnn}). The training process, requiring about 40 min for each CNN, was repeated 10 times and resulted in the following average accuracy: 99.4$\pm$0.4 \% for B\_V CNN, 99.6$\pm$0.3\% for H\_E CNN and 97.9$\pm$0.6\% for E\_I CNN.  The confusion matrices of one of the validation processes are reported in Fig~\ref{fig:matrix}. They confirmed that the class E\_I was the most difficult to be recognized.

\begin{table}[t]
	\begin{center}
	\resizebox{\textwidth}{!}{
		\begin{tabular}{lc|c|c|c|c|}
			\cline{3-6}
			\multicolumn{2}{l|}{\multirow{2}{*}{}}                            & \multicolumn{4}{c|}{\textbf{Classification Results}}                                               \\ \cline{3-6} 
			\multicolumn{2}{l|}{}                                             & \textbf{Others}            & \textbf{Artifacts}     & \multicolumn{2}{c|}{\textbf{Double Detections}} \\ \cline{1-6} 
			\multicolumn{1}{|l|}{\textbf{}}  & \multirow{2}{*}{\textbf{B\_V}} & \multirow{2}{*}{310720} & \multirow{2}{*}{30120}  & \textbf{H\_E}          & \textbf{E\_I}          \\ \cline{5-6} 
			\multicolumn{1}{|c|}{\textbf{}} &                                &                         &                        & 340                     & 4480                     \\ \cline{2-6} 
			\multicolumn{1}{|c|}{\textbf{}} & \multirow{2}{*}{\textbf{H\_E}} & \multirow{2}{*}{332429} & \multirow{2}{*}{8411} & \textbf{B\_V}          & \textbf{E\_I}          \\ \cline{5-6} 
			\multicolumn{1}{|c|}{\textbf{Framework}} &                                &                         &                        & 340                     & 320                    \\ \cline{2-6} 
			\multicolumn{1}{|c|}{\textbf{}} & \multirow{2}{*}{\textbf{E\_I}} & \multirow{2}{*}{296669} & \multirow{2}{*}{44171} & \textbf{B\_V}          & \textbf{H\_E}          \\ \cline{5-6} 
			\multicolumn{1}{|l|}{}           &                                &                         &                        & 4480                    & 320                   \\ \cline{1-6} 
	\end{tabular}
	}
	\end{center}
	\caption{Classification results on 340890 topoplots. The column "Others" contains other artifacts + UBS. The column "Artifacts" contains the number of artifacts recognized as proper by the CNN indicated on the left. Reciprocal double detections are artifacts considered as proper by two CNNs.}
	\label{tab:table2}
\end{table}

%\begin{figure}[t]
%\begin{center}
%%\fbox{\rule{0pt}{2in} \rule{0.9\linewidth}{0pt}}
%   \includegraphics[width=0.9\linewidth]{images/pre_processing_time.pdf}
%\end{center}
%   \caption{Time course of the pre-processing pipeline: 2.3 sec were necessary to evaluate 64 component topoplots allowing to a temporal slice signal of 2.5 sec.}
%\label{fig:timing_performances}
%\end{figure}
%

%\begin{figure*}
%\begin{center}
%%\fbox{\rule{0pt}{2in} \rule{0.9\linewidth}{0pt}}
%   \includegraphics[width=0.9\linewidth]{images/training_graphics_cnn.png}
%\end{center}
%   \caption{Configuration and disposition of the 3 CNNs inside the classification system. The topoplot to be classified is passed to the three CNN separately.}
%\label{fig:aaa}
%\end{figure*}

\subsection{Results and discussion}

The proposed framework was tested on data from subjects 20-32 (DEAP dataset) corresponding to 340890 images. Classification results, summarized in Table \ref{tab:table2}, show that, between the topoplots considered as artifacts, some (very few) were shared between two different classes of artifacts and, to a deep verification with the consensus of the human experts, most of them effectively showed ambiguous patterns (equity, see above). No topoplot was shared between the three classes at the same time. 

\begin{figure*}[h]
	\begin{center}
		\includegraphics[width=\textwidth]{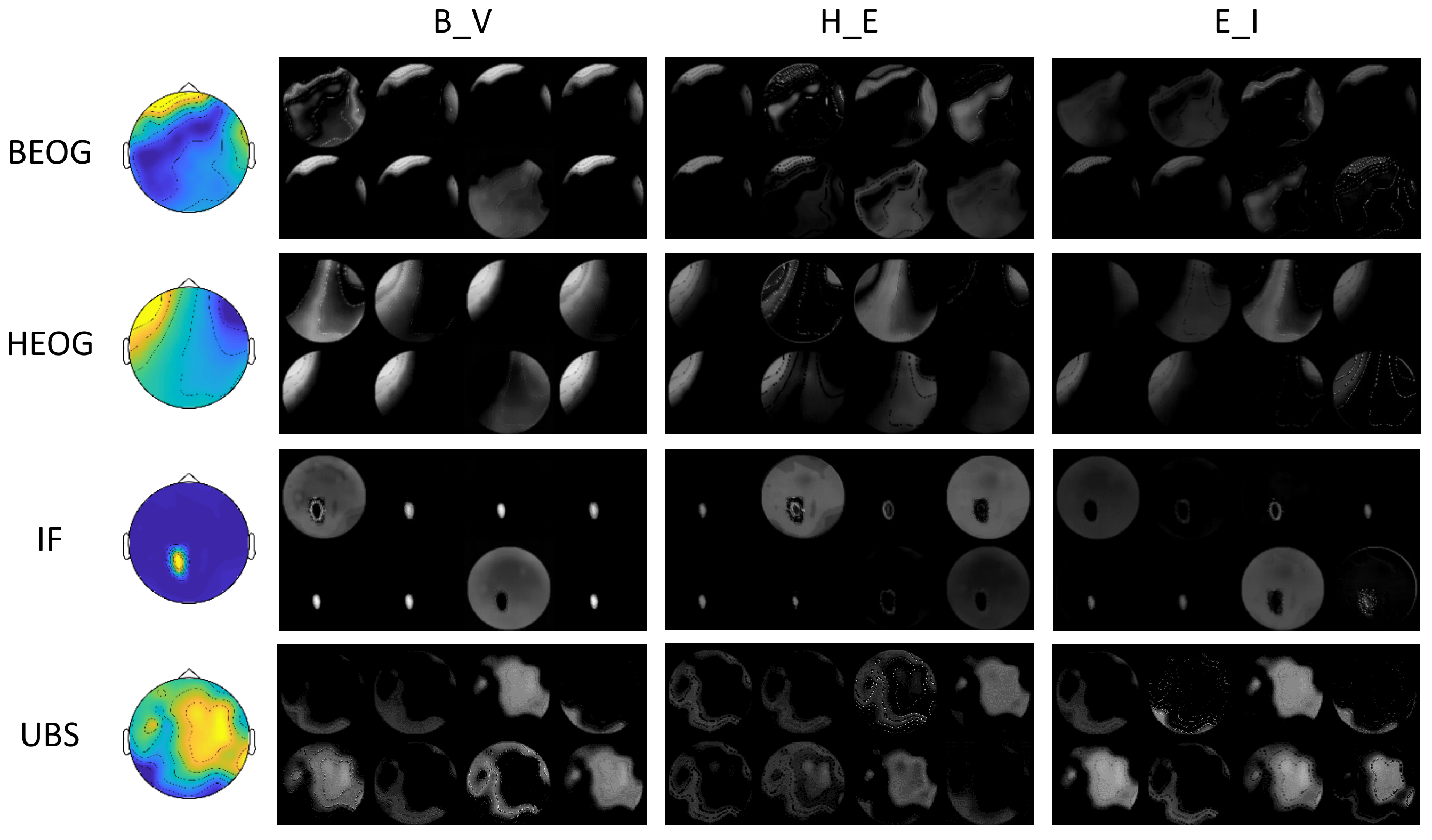}
	\end{center}
	\caption{First feature extraction step of the three CNNs (columns) when receiving in input the topoplot of each of the three considered classes of artifacts or are UBS (rows). The features extracted in the following steps are not shown.}
	\label{fig:FeatureExtraction}
\end{figure*}

The behaviour of each CNN with respect to the others, when acting on different types of topoplots, is shown in Fig. \ref{fig:FeatureExtraction}. In particular, the output of the first feature extraction step for each CNN, when acting on the same topoplot (rows) or repeated for different topoplots (columns), are presented. As can be observed, though at the first level of feature extraction, the CNNs reacted very differently to the same topoplot, thus demonstrating a good fixation activity \cite{mopuri2018cnn} of the CNNs. The topoplots reported in Fig. \ref{fig:FeatureExtraction} represents examples of all classes of artifacts and UBS. 

\begin{figure*}[h]
	\begin{center}
		\includegraphics[width=\textwidth]{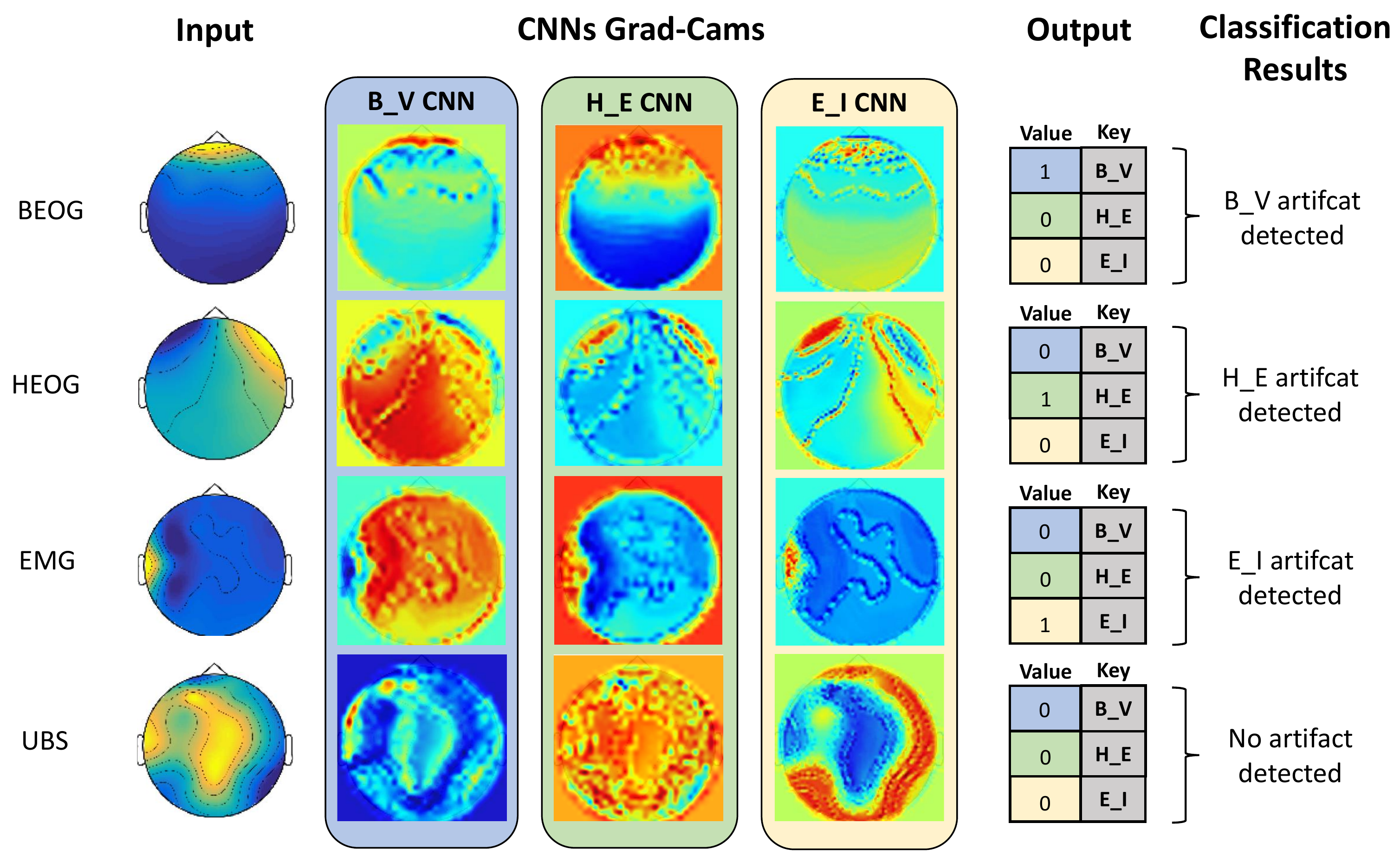}
	\end{center}
	\caption{Grad-Cams of the three CNNs (columns) when receiving in input the topoplot of each of the three considered classes of artifacts or are UBS(rows). The last two columns on the right show the Output and the Classification results, respectively.}
	\label{fig:gradCam}
\end{figure*}

Moreover we used Gradient-weighted Class Activation Mapping (Grad-Cam) to produce a coarse localization map highlighting the important regions in the image used for prediction \cite{selvaraju2017grad}. Fig. \ref{fig:gradCam} shows examples of grad-cams organized in a table: rows report the 4 different classes of topolots (artifacts + UBS) and columns represent the respective grad-cams of each CNNs. In the first row a B\_V artifact is passed through the CNNs. The B\_V CNN grad-cam shows that the activations are correctly localized on the frontal region of the scalp. In the second row a H\_E artifact is passed through the CNNs and the H\_E CNN grad-cam shows that the activations are corrected localized on the frontal region of the scalp: this confirms that the H\_E CNN is not biased by the strongest positive activation (yellow) in the topolot. In the third row, an E\_I artifact is passed through the CNNs and the E\_I CNN grad-cam shows that the positive values are well localized on the impedance artifact while the rest of scalp is negative (this suggests that the E\_I CNN checks the whole topoplot).
Finally, in the last row an UBS topolot is passed through the CNNs and none of them recognizes it as allowing to its proper class.

\begin{figure}[h]
	\begin{center}
		\includegraphics[width=0.7\textwidth]{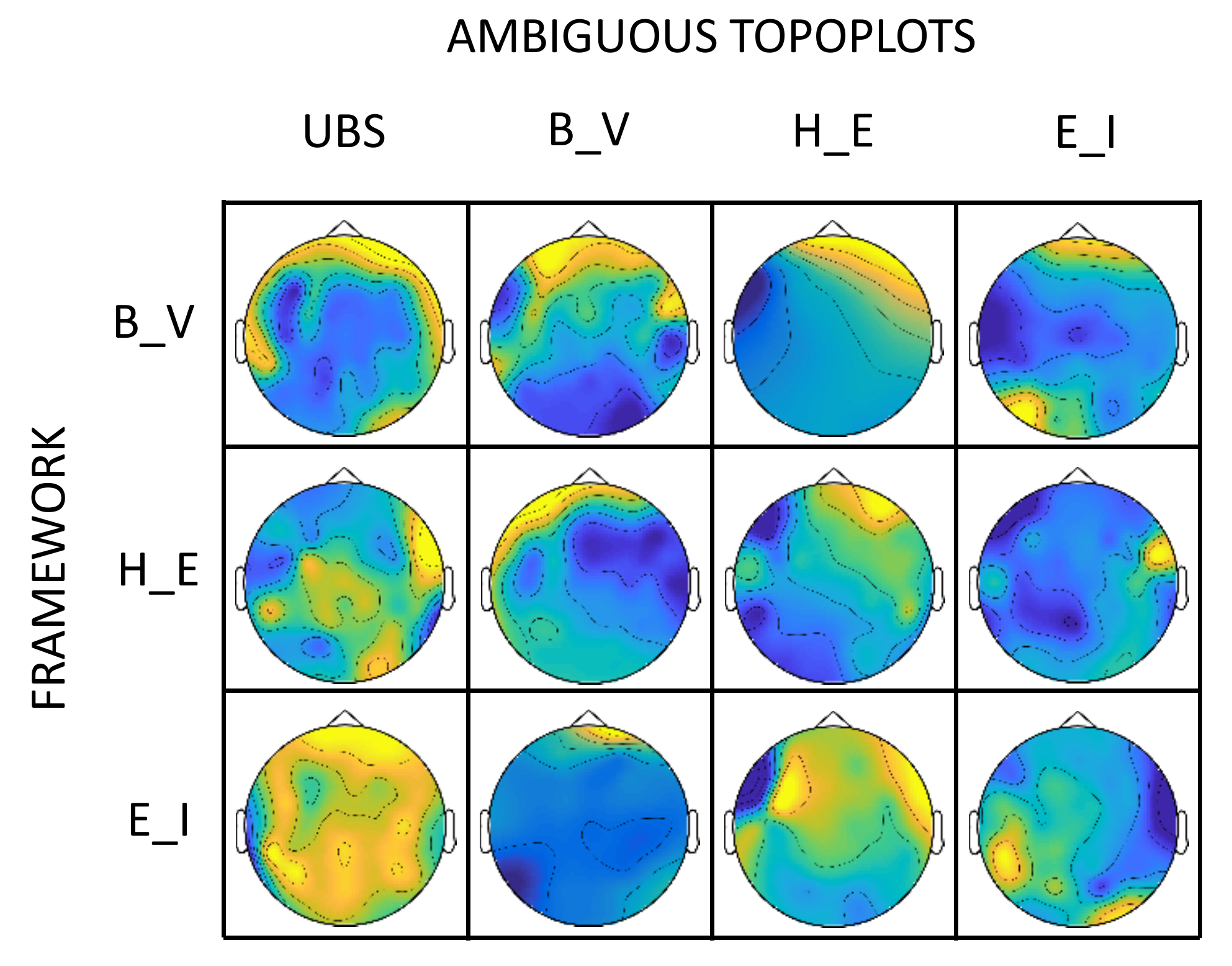}
	\end{center}
	\caption{Examples of ambiguous topoplots. The figure has the same meaning of Table \ref{tab:table3} (excluding "performance") with topoplots instead of numbers.}
	\label{fig:aaa}
\end{figure}

To check deeply the behaviour of the framework, another dataset of topoplots from subjects 9-19 was generated and 1/10 of them (a total of 29500), randomly selected, was  submitted to the 5 human experts for visual classification and labelling. The consensus contained 22795 UBS, 2190 B\_V, 526 H\_E and 3871 E\_I, respectively. A set of 117 topoplots was discarded because of equity (see above). Then, automatic recognition was performed by using the proposed framework: the resulting errors, with respect to manual consensus, and the performance obtained by the three CNNs are reported in Table \ref{tab:table3}. 
Performance values demonstrate that our framework shows very good accuracy, sensitivity and specificity ($>$98\%). Moreover, we verified that different humans experts disagreed on most of the topoplots wrongly classified by the framework, thus confirming that most of them showed ambiguous (mixed) pattern. Examples of ambiguous topoplots are reported in Fig \ref{fig:aaa}.

Finally, we made an indirect comparison between out framework and that proposed by Pion-Tonachini et al~\cite{Pion2019} by asking to our 5 human experts to instruct themselves, by using the on-line tools of the web-site (INSERIRE SITO), on how to use, besides topoplots, PSD and autocorrelation to classify ICs. Then, 1000 topoplot images of artifacts (both at 32x32 and at 134x136 pixels) were generated, including PSD and autocorrelation and to the experts was asked to classify them both with the classic visual inspection of just the high resolution topoplots and with the method learned on the web-site, that uses both the 32x32 topoplots and the other information. Double consensus were generated: they agreed at 95.4\%, though the time spent for classic classification was about 1/5 of that necessary for using additional information. The strong agreement with the consensus demonstrates that the introduction of further information could contribute just for a few percentage to the classification improvement (in line with the disagreement between experts, that remained almost the same also by using additional information), but it produces a strong increment of the time necessary for manual classification. Moreover, though the proposed framework could be less precise of the strategy in~\cite{Pion2019} (a complete comparison of both automatic strategies should be carried on), its usage on fast-response EEG is completely justified because the inspection of just high resolution topoplots, is effective in defining the nature of the ICs, as usually done by human experts.

\begin{table}[t]
	\begin{center}
		\resizebox{\textwidth}{!}{%
			\begin{tabular}{cc|c|c|c|c|c?c|c|c|}
				\cline{3-10}
				\multicolumn{2}{c|}{\multirow{2}{*}{\textbf{}}}                       & \multicolumn{5}{c?}{\textbf{CLASSIFICATION ERRORS (number)}}                                                 & \multicolumn{3}{c|}{\textbf{PERFORMANCE (\%)}}                                                                                     \\ \cline{3-10} 
				\multicolumn{2}{c|}{}                                                 & \textbf{UBS} & \textbf{B\_V} & \textbf{H\_E} & \textbf{E\_I} & \textbf{TOTAL}  & \multicolumn{1}{l|}{\textbf{ACC.}} & \multicolumn{1}{l|}{\textbf{SENS.}} & \multicolumn{1}{l|}{\textbf{SPEC.}} \\ \hline
				\multicolumn{1}{|c|}{\multirow{3}{*}{\textbf{Framework}}} & \textbf{B\_V} & 346                  & 20 (FN)                 & 17                   & 235                   & 618            & 98.7                                 & 99.0                                    & 98.7 \\ \cline{2-10} 
				\multicolumn{1}{|c|}{}                               & \textbf{H\_E} & 114                  & 16                   & 9 (FN)       & 22                  & 161            & 99.5                                  & 98.2                                     & 99.6                                    \\ \cline{2-10} 
				\multicolumn{1}{|c|}{}                               & \textbf{E\_I} & 162                 & 134                   & 138 & 73 (FN) & 507 & 99.1                               & 98.1                                     & 99.2                                    \\ \hline
				%\multicolumn{2}{|c|}{\textbf{The Framework}}                                & 220                 & \multicolumn{3}{c|}{95}                                        %& 315           & 98.9                                & 98.5                                     & 99.0                                    %\\ \hline
		\end{tabular}}
	\end{center}
	\caption{Number of recognition errors for each CNN when acting on a labeled dataset of topoplots generated by subjects 9-19. The column "TOTAL" contains the sum of the values to its left. The number of false negatives (indicated with FN close to the number) are into cells where the same name occurs on crossing row and column. The others are false positive. Last three columns report the performance (accuracy, sensitivity and specificity) in percentage.}
	\label{tab:table3}
\end{table}

In terms of computational performance, the proposed automatic framework required 1.4 sec to classify 32 topoplots, including: time required for ICA (0.3 sec by using the fast-ICA method proposed in \cite{Hsu2016}); time  necessary to generate the topoplots (0.9 sec); time required for classification (0.21 sec).
As can be seen, the bottleneck is represented by topoplots generation (this time was also necessary both for manual inspection and for other automatic strategies): fortunately, this time does not increase linearly with the number of images because some of the calculations, such as those required for channels positioning on the scalp, are executed just once. Despite all, our framework performed sufficiently fast to be compatible with fast-response EEG. In fact, the proposed method is simpler and, probably, more efficient than the strategy in ~\cite{Pion2019} due to the fact that this last method, besides topoplots, has to analyse also other information. This is particularly important because it demonstrates that pre-processing can be carried on with the granularity of a sub-trial, thus making artifacts recognition effective and timely. This makes automatic IC topoplot processing available also for fast BCI applications \cite{jalilpour2020novel,chai2020effects, khare2020facile}.

\section{Conclusion}

The use of an automatic AI strategy for artifacts recognition from IC topoplots is fundamental, since it is considered the gold standard for this purpose. We have demonstrated that: its feasibility with a CNN-based framework is possible just operating on IC topoplots, without the need of using additional information; it can be completely automatic; it can be performed with high accuracy ($\sim$99\%), specificity ($\sim$98\%) and sensitivity ($\sim$99\%), with about no differences with human experts; its results are in line with those of other, more complex methods ~\cite{Pion2019};
it can be applicable in BCIs based on EEG. Our framework, thanks to its scalable structure, greatly fits present and future requirements of artifacts recognition because it can be easily adapted and trained to deal with newly discovered future artifact patterns without re-train the existing architecture (specific CNNs have to be added to the framework and trained separately).

The proposed framework was capable to operate in 1.4 sec for 32 components topoplots (including time necessary to generate topoplots), fast with respect to the time necessary to collect EEG signals for an EEG-based BCI (comprised between 2.5-5 sec). Moreover time can be reduced if a lower number of sensors is used (as usual in BCI) without any framework's modification.  
Though the time spent for classification is low (0.2 sec), two considerations are important: 1) some of the operations necessary in the pre-processing stage (such as ICA) are also necessary in the following processing stage; 2) the pre-processing computation time was obtained with scripts written in Matlab, without CPU multithreading or GPU acceleration. By speeding up the topoplot generation and by rewriting the software in a compiled language (such as C/C++), we could considerably improve its efficiency. 
However, to further improve the system efficiency, work is in progress along three directions: 1) to demonstrate the equivalence between gray-scale totoplots and RGB topoplots (for a human RGB is necessary) in order to reduce by 3 the number of analyzed images; 2) to optimize the networks architecture through automatic strategies~\cite{shahriari2015taking}; 3) to directly compare the proposed framework with the method in~\cite{Pion2019} in terms of performance, simplicity and generalizability. Work is also in progress to test the framework into an emotion-based BCI used for minimally conscious people~\cite{Pistoia2015} whose UBS are weak (brain signals generated in deep, primordial, regions of the brain) and artifacts frequently occurs (muscular spasms or uncontrolled ocular movements are frequent). 

%\bibliographystyle{model2-names}
%\bibliography{refs}

%----------------------------------------------------------------------------------------
%	BIBLIOGRAPHY
%----------------------------------------------------------------------------------------

\renewcommand{\refname}{\spacedlowsmallcaps{References}} % For modifying the bibliography heading

\bibliographystyle{unsrt}

\bibliography{refs}

\begin{thebibliography}{10}

\bibitem{suk2012novel}
Heung-Il Suk and Seong-Whan Lee.
\newblock A novel bayesian framework for discriminative feature extraction in
  brain-computer interfaces.
\newblock {\em IEEE Transactions on Pattern Analysis and Machine Intelligence},
  35(2):286--299, 2012.

\bibitem{cecotti2010convolutional}
Hubert Cecotti and Axel Graser.
\newblock Convolutional neural networks for p300 detection with application to
  brain-computer interfaces.
\newblock {\em IEEE transactions on pattern analysis and machine intelligence},
  33(3):433--445, 2010.

\bibitem{Placidi2015}
Giuseppe Placidi, Danilo Avola, Andrea Petracca, Fiorella Sgallari, and Matteo
  Spezialetti.
\newblock {Basis for the implementation of an EEG-based single-trial binary
  brain computer interface through the disgust produced by remembering
  unpleasant odors}.
\newblock {\em Neurocomputing}, 160:308--318, jul 2015.

\bibitem{Uriguen2015}
Jose~Antonio Urig{\"{u}}en and Bego{\~{n}}a Garcia-Zapirain.
\newblock {EEG artifact removal—state-of-the-art and guidelines}.
\newblock {\em Journal of Neural Engineering}, 12(3):031001, jun 2015.

\bibitem{noureddin2011online}
Borna Noureddin, Peter~D Lawrence, and Gary~E Birch.
\newblock Online removal of eye movement and blink eeg artifacts using a
  high-speed eye tracker.
\newblock {\em IEEE Transactions on Biomedical Engineering}, 59(8):2103--2110,
  2011.

\bibitem{Joyce2004}
Carrie~A. Joyce, Irina~F. Gorodnitsky, and Marta Kutas.
\newblock {Automatic removal of eye movement and blink artifacts from EEG data
  using blind component separation}.
\newblock {\em Psychophysiology}, 41(2):313--325, mar 2004.

\bibitem{Lin2014}
P.-F. Lin, M.-T. Lo, J.~Tsao, Y.-C. Chang, C.~Lin, and Y.-L. Ho.
\newblock Correlations between the signal complexity of cerebral and cardiac
  electrical activity: A multiscale entropy analysis.
\newblock {\em PLoS ONE}, 9(2), 2014.

\bibitem{Shibas1999}
H.~Shibasaki and J.C. Rothwell.
\newblock Emg-eeg correlation. the international federation of clinical
  neurophysiology.
\newblock {\em Electroencephalography and clinical neurophysiology.
  Supplement}, 52:269--274, 1999.

\bibitem{Mognon2011}
Andrea Mognon, Jorge Jovicich, Lorenzo Bruzzone, and Marco Buiatti.
\newblock {ADJUST: An automatic EEG artifact detector based on the joint use of
  spatial and temporal features}.
\newblock {\em Psychophysiology}, 48(2):229--240, feb 2011.

\bibitem{Delorme2007}
Arnaud Delorme, Terrence Sejnowski, and Scott Makeig.
\newblock {Enhanced detection of artifacts in EEG data using higher-order
  statistics and independent component analysis}.
\newblock {\em NeuroImage}, 34(4):1443--1449, feb 2007.

\bibitem{zhang2019making}
Dalin Zhang, Lina Yao, Kaixuan Chen, Sen Wang, Xiaojun Chang, and Yunhao Liu.
\newblock Making sense of spatio-temporal preserving representations for
  eeg-based human intention recognition.
\newblock {\em IEEE transactions on cybernetics}, 2019.

\bibitem{Jung2000163}
T.-P. Jung, S.~Makeig, C.~Humphries, T.-W. Lee, M.J. Mckeown, V.~Iragui, and
  T.J. Sejnowski.
\newblock Removing electroencephalographic artifacts by blind source
  separation.
\newblock {\em Psychophysiology}, 37(2):163--178, 2000.

\bibitem{Raduntz2015}
T.~Rad{\"{u}}ntz, J.~Scouten, O.~Hochmuth, and B.~Meffert.
\newblock {EEG artifact elimination by extraction of ICA-component features
  using image processing algorithms}.
\newblock {\em Journal of Neuroscience Methods}, 243:84--93, mar 2015.

\bibitem{Delorme2004}
Arnaud Delorme and Scott Makeig.
\newblock {EEGLAB: an open source toolbox for analysis of single-trial EEG
  dynamics including independent component analysis}.
\newblock {\em Journal of Neuroscience Methods}, 134(1):9--21, mar 2004.

\bibitem{Oostenveld2011}
Robert Oostenveld, Pascal Fries, Eric Maris, and Jan-Mathijs Schoffelen.
\newblock {FieldTrip: Open Source Software for Advanced Analysis of MEG, EEG,
  and Invasive Electrophysiological Data}.
\newblock {\em Computational Intelligence and Neuroscience}, 2011:1--9, 2011.

\bibitem{Koelstra2011deap}
Sander Koelstra, Christian Muhl, Mohammad Soleymani, Jong-Seok Lee, Ashkan
  Yazdani, Touradj Ebrahimi, Thierry Pun, Anton Nijholt, and Ioannis Patras.
\newblock Deap: A database for emotion analysis; using physiological signals.
\newblock {\em IEEE transactions on affective computing}, 3(1):18--31, 2011.

\bibitem{Islam2016}
Md~Kafiul Islam, Amir Rastegarnia, and Zhi Yang.
\newblock {Methods for artifact detection and removal from scalp EEG: A
  review}.
\newblock {\em Neurophysiologie Clinique/Clinical Neurophysiology},
  46(4-5):287--305, nov 2016.

\bibitem{Mannan2018}
Malik Muhammad~Naeem Mannan, Muhammad~Ahmad Kamran, and Myung~Yung Jeong.
\newblock {Identification and Removal of Physiological Artifacts From
  Electroencephalogram Signals: A Review}.
\newblock {\em IEEE Access}, 6:30630--30652, 2018.

\bibitem{VandenBerg-Lenssen1994}
M.~M.~C. van~den Berg-Lenssen, J.~A.~M. van Gisbergen, and B.~W. Jervis.
\newblock {Comparison of two methods for correcting ocular artefacts in EEGs}.
\newblock {\em Medical and Biological Engineering and Computing},
  32(5):501--511, sep 1994.

\bibitem{Minguillon2017}
Jesus Minguillon, M.~Angel Lopez-Gordo, and Francisco Pelayo.
\newblock {Trends in EEG-BCI for daily-life: Requirements for artifact
  removal}.
\newblock {\em Biomedical Signal Processing and Control}, 31:407--418, jan
  2017.

\bibitem{Romero2008}
Sergio Romero, Miguel~A. Ma{\~{n}}anas, and Manel~J. Barbanoj.
\newblock {A comparative study of automatic techniques for ocular artifact
  reduction in spontaneous EEG signals based on clinical target variables: A
  simulation case}.
\newblock {\em Computers in Biology and Medicine}, 38(3):348--360, mar 2008.

\bibitem{Sweeney2012}
K.~T. Sweeney, T.~E. Ward, and S.~F. McLoone.
\newblock {Artifact Removal in Physiological Signals—Practices and
  Possibilities}.
\newblock {\em IEEE Transactions on Information Technology in Biomedicine},
  16(3):488--500, may 2012.

\bibitem{acharyya2018low}
Amit Acharyya, Pranit~N Jadhav, Valentina Bono, Koushik Maharatna, and Ganesh~R
  Naik.
\newblock Low-complexity hardware design methodology for reliable and automated
  removal of ocular and muscular artifact from eeg.
\newblock {\em Computer methods and programs in biomedicine}, 158:123--133,
  2018.

\bibitem{Berg1991}
P~Berg and M~Scherg.
\newblock {Dipole modelling of eye activity and its application to the removal
  of eye artefacts from the EEG and MEG}.
\newblock {\em Clinical Physics and Physiological Measurement}, 12(A):49--54,
  jan 1991.

\bibitem{Lagerlund1997}
Terrence~D. Lagerlund, Frank~W. Sharbrough, and Neil~E. Busacker.
\newblock {Spatial Filtering of Multichannel Electroencephalographic Recordings
  Through Principal Component Analysis by Singular Value Decomposition}.
\newblock {\em Journal of Clinical Neurophysiology}, 14(1):73--82, jan 1997.

\bibitem{Zhao2014109}
Q.~Zhao, B.~Hu, Y.~Shi, Y.~Li, P.~Moore, M.~Sun, and H.~Peng.
\newblock Automatic identification and removal of ocular artifacts in eeg -
  improved adaptive predictor filtering for portable applications.
\newblock {\em IEEE Transactions on Nanobioscience}, 13(2):109--117, 2014.

\bibitem{Urrestarazu2004}
Elena Urrestarazu, Jorge Iriarte, Manuel Alegre, Miguel Valencia, C{\'{e}}sar
  Viteri, and Julio Artieda.
\newblock {Independent Component Analysis Removing Artifacts in Ictal
  Recordings}.
\newblock {\em Epilepsia}, 45(9):1071--1078, sep 2004.

\bibitem{Vigario2000}
R.~Vigario, J.~Sarela, V.~Jousmiki, M.~Hamalainen, and E.~Oja.
\newblock {Independent component approach to the analysis of EEG and MEG
  recordings}.
\newblock {\em IEEE Transactions on Biomedical Engineering}, 47(5):589--593,
  may 2000.

\bibitem{Raduntz2017}
Thea Rad{\"{u}}ntz, Jon Scouten, Olaf Hochmuth, and Beate Meffert.
\newblock {Automated EEG artifact elimination by applying machine learning
  algorithms to ICA-based features}.
\newblock {\em Journal of Neural Engineering}, 14(4):046004, aug 2017.

\bibitem{James2003}
C.J. James and O.J. Gibson.
\newblock {Temporally constrained ica: an application to artifact rejection in
  electromagnetic brain signal analysis}.
\newblock {\em IEEE Transactions on Biomedical Engineering}, 50(9):1108--1116,
  sep 2003.

\bibitem{Winkler2014}
Irene Winkler, Stephanie Brandl, Franziska Horn, Eric Waldburger, Carsten
  Allefeld, and Michael Tangermann.
\newblock {Robust artifactual independent component classification for BCI
  practitioners}.
\newblock {\em Journal of Neural Engineering}, 11(3):035013, jun 2014.

\bibitem{Winkler2014a}
Irene Winkler, Stephanie Brandl, Franziska Horn, Eric Waldburger, Carsten
  Allefeld, and Michael Tangermann.
\newblock {Robust artifactual independent component classification for BCI
  practitioners}.
\newblock {\em Journal of Neural Engineering}, 11(3):035013, jun 2014.

\bibitem{Pion2019}
Luca Pion-Tonachini, Ken Kreutz-Delgado, and Scott Makeig.
\newblock Iclabel: An automated electroencephalographic independent component
  classifier, dataset, and website.
\newblock {\em NeuroImage}, 198:181 -- 197, 2019.

\bibitem{al2017review}
Ahmed Ali~Mohammed Al-Saffar, Hai Tao, and Mohammed~Ahmed Talab.
\newblock Review of deep convolution neural network in image classification.
\newblock In {\em 2017 International Conference on Radar, Antenna, Microwave,
  Electronics, and Telecommunications (ICRAMET)}, pages 26--31. IEEE, 2017.

\bibitem{song2018deep}
Lingyun Song, Jun Liu, Buyue Qian, Mingxuan Sun, Kuan Yang, Meng Sun, and Samar
  Abbas.
\newblock A deep multi-modal cnn for multi-instance multi-label image
  classification.
\newblock {\em IEEE Transactions on Image Processing}, 27(12):6025--6038, 2018.

\bibitem{askari2018modeling}
Elham Askari, Seyed~Kamaledin Setarehdan, Ali Sheikhani, Mohammad~Reza
  Mohammadi, and Mohammad Teshnehlab.
\newblock Modeling the connections of brain regions in children with autism
  using cellular neural networks and electroencephalography analysis.
\newblock {\em Artificial intelligence in medicine}, 89:40--50, 2018.

\bibitem{tang2019motor}
Xianlun Tang, Ting Wang, Yiming Du, and Yuyan Dai.
\newblock Motor imagery eeg recognition with knn-based smooth auto-encoder.
\newblock {\em Artificial Intelligence in Medicine}, 101:101747, 2019.

\bibitem{gao2020automatic}
Xiaozeng Gao, Xiaoyan Yan, Ping Gao, Xiujiang Gao, and Shubo Zhang.
\newblock Automatic detection of epileptic seizure based on approximate
  entropy, recurrence quantification analysis and convolutional neural
  networks.
\newblock {\em Artificial Intelligence in Medicine}, 102:101711, 2020.

\bibitem{Lecun1989}
Yann LeCun, Bernhard Boser, John~S Denker, Donnie Henderson, Richard~E Howard,
  Wayne Hubbard, and Lawrence~D Jackel.
\newblock Backpropagation applied to handwritten zip code recognition.
\newblock {\em Neural computation}, 1(4):541--551, 1989.

\bibitem{lecun1998gradient}
Yann LeCun, L{\'e}on Bottou, Yoshua Bengio, Patrick Haffner, et~al.
\newblock Gradient-based learning applied to document recognition.
\newblock {\em Proceedings of the IEEE}, 86(11):2278--2324, 1998.

\bibitem{nuwer1998ifcn}
Marc~R Nuwer, Giancarlo Comi, Ronald Emerson, Anders Fuglsang-Frederiksen,
  Jean-Michel Gu{\'e}rit, Hermann Hinrichs, Akio Ikeda, Fransisco Jose~C
  Luccas, and Peter Rappelsburger.
\newblock Ifcn standards for digital recording of clinical eeg.
\newblock {\em Electroencephalography and clinical Neurophysiology},
  106(3):259--261, 1998.

\bibitem{Leske2019}
Sabine Leske and Sarang~S. Dalal.
\newblock Reducing power line noise in eeg and meg data via spectrum
  interpolation.
\newblock {\em NeuroImage}, 189:763 -- 776, 2019.

\bibitem{liu2018method}
Zhipeng Liu, Lichun Li, Haiyun Xu, and Huiqi Li.
\newblock A method for recognition and classification for hybrid signals based
  on deep convolutional neural network.
\newblock In {\em 2018 International Conference on Electronics Technology
  (ICET)}, pages 325--330. IEEE, 2018.

\bibitem{ito2018application}
Chihiro Ito, Xin Cao, Masaki Shuzo, and Eisaku Maeda.
\newblock Application of cnn for human activity recognition with fft
  spectrogram of acceleration and gyro sensors.
\newblock In {\em Proceedings of the 2018 ACM International Joint Conference
  and 2018 International Symposium on Pervasive and Ubiquitous Computing and
  Wearable Computers}, pages 1503--1510. ACM, 2018.

\bibitem{siddharth2019utilizing}
Siddharth Siddharth, Tzyy-Ping Jung, and Terrence~J Sejnowski.
\newblock Utilizing deep learning towards multi-modal bio-sensing and
  vision-based affective computing.
\newblock {\em IEEE Transactions on Affective Computing}, 2019.

\bibitem{mopuri2018cnn}
Konda~Reddy Mopuri, Utsav Garg, and R~Venkatesh Babu.
\newblock Cnn fixations: an unraveling approach to visualize the discriminative
  image regions.
\newblock {\em IEEE Transactions on Image Processing}, 28(5):2116--2125, 2018.

\bibitem{selvaraju2017grad}
Ramprasaath~R Selvaraju, Michael Cogswell, Abhishek Das, Ramakrishna Vedantam,
  Devi Parikh, and Dhruv Batra.
\newblock Grad-cam: Visual explanations from deep networks via gradient-based
  localization.
\newblock In {\em Proceedings of the IEEE international conference on computer
  vision}, pages 618--626, 2017.

\bibitem{Hsu2016}
Sheng-Hsiou Hsu, Tim~R. Mullen, Tzyy-Ping Jung, and Gert Cauwenberghs.
\newblock {Real-Time Adaptive EEG Source Separation Using Online Recursive
  Independent Component Analysis}.
\newblock {\em IEEE Transactions on Neural Systems and Rehabilitation
  Engineering}, 24(3):309--319, mar 2016.

\bibitem{jalilpour2020novel}
Shayan Jalilpour, Sepideh~Hajipour Sardouie, and Amirmohammad Mijani.
\newblock A novel hybrid bci speller based on rsvp and ssvep paradigm.
\newblock {\em Computer Methods and Programs in Biomedicine}, 187:105326, 2020.

\bibitem{chai2020effects}
Xiaoke Chai, Zhimin Zhang, Kai Guan, Tengyu Zhang, Jinxiu Xu, and Haijun Niu.
\newblock Effects of fatigue on steady state motion visual evoked potentials:
  Optimised stimulus parameters for a zoom motion-based brain-computer
  interface.
\newblock {\em Computer Methods and Programs in Biomedicine}, page 105650,
  2020.

\bibitem{khare2020facile}
Smith~K Khare and Varun Bajaj.
\newblock A facile and flexible motor imagery classification using
  electroencephalogram signals.
\newblock {\em Computer Methods and Programs in Biomedicine}, page 105722,
  2020.

\bibitem{shahriari2015taking}
Bobak Shahriari, Kevin Swersky, Ziyu Wang, Ryan~P Adams, and Nando De~Freitas.
\newblock Taking the human out of the loop: A review of bayesian optimization.
\newblock {\em Proceedings of the IEEE}, 104(1):148--175, 2015.

\bibitem{Pistoia2015}
F.~Pistoia, A.~Carolei, D.~Iacoviello, A.~Petracca, S.~Sacco, M.~Sarà,
  M.~Spezialetti, and G.~Placidi.
\newblock Eeg-detected olfactory imagery to reveal covert consciousness in
  minimally conscious state.
\newblock {\em Brain Injury}, 29(13-14):1729--1735, 2015.

\end{thebibliography}

%----------------------------------------------------------------------------------------

\end{document}